\documentclass[final]{cvpr}

\usepackage{times}
\usepackage{epsfig}
\usepackage{graphicx}
\usepackage{amsmath}
\usepackage{amssymb}
\usepackage[titlenumbered,ruled,vlined]{algorithm2e}
\usepackage{ccaption}
\newfixedcaption{\outfigcaption}{figure}
\usepackage{comment}
\usepackage{kotex}
\usepackage{color}
\usepackage{url}
\usepackage{appendix}

\makeatletter
\@namedef{ver@everyshi.sty}{}
\makeatother
\usepackage{pgfplots}
\pgfplotsset{compat=1.16}

\usepackage{multirow}

\usepackage[pagebackref=true,breaklinks=true,colorlinks,bookmarks=false]{hyperref}

\def\cvprPaperID{4816} 
\def\confYear{CVPR 2021}

\begin{document}

\newcommand{\ignore}[1]{}
\renewcommand{\tt}{\mathbf{t}} 
\newcommand{\tp}{\mathbf{t}} 
\newcommand{\tw}{\mathbf{w}} 
\newcommand{\ptp}{\mathbf{pt}} 
\newcommand{\tc}{\mathbf{c}} 
\newcommand{\tu}{\mathbf{u}} 
\newcommand{\tb}{\mathbf{b}} 
\newcommand{\ta}{\mathbf{a}} 
\newcommand{\te}{\mathbf{e}}
\newcommand{\PT}{\mathsf{Protect}} 
\newcommand{\VF}{\mathsf{Verify}}
\newcommand{\M}{\mathcal{M}} 
\newcommand{\helper}{\mathbf{P}} 
\newcommand{\dist}{\mathsf{dist}} 
\newcommand{\TPTS}{\textrm{IronMask}} 
\newcommand{\hash}{\mathsf{H}} 

\renewcommand{\th}{\alpha}

\newcommand{\HT}{\textcolor{blue}{HT}}


\title{IronMask: Modular Architecture for Protecting Deep Face Template}

\author{$\text{Sunpill Kim}^{1}$ \ $\text{Yunseong Jeong}^{2}$ \ $\text{Jinsu Kim}^{3}$ \ $\text{Jungkon Kim}^{3}$ \ $\text{Hyung Tae Lee}^{2}$ \ $\text{Jae Hong Seo}^{1}$\thanks{Corresponding author: J. H. Seo (e-mail: jaehongseo@hanyang.ac.kr)}\\ \\
$^{1}\text{Department of Mathematics \& Research Institute for Natural Sciences, Hanyang University}$ \\
$^{2}\text{Division of Computer Science and Engineering, College of Engineering, Jeonbuk National University}$\\
$^{3}\text{Security Team, Samsung Research, Samsung Electronics}$
}

\maketitle
\thispagestyle{empty}

\begin{abstract}
Convolutional neural networks have made remarkable progress in the face recognition field.
The more the technology of face recognition advances, the greater discriminative features into a face template. However, this increases the threat to user privacy in case the template is exposed.

In this paper, we present a modular architecture for face template protection, called IronMask, that can be combined with any face recognition system using angular distance metric. We circumvent the need for binarization, which is the main cause of performance degradation in most existing face template protections, by proposing a new real-valued error-correcting-code that is compatible with real-valued templates and can therefore, minimize performance degradation. We evaluate the efficacy of IronMask by extensive experiments on two face recognitions, ArcFace and CosFace with three datasets, CMU-Multi-PIE, FEI, and Color-FERET. According to our experimental results, IronMask achieves a true accept rate (TAR) of 99.79\% at a false accept rate (FAR) of 0.0005\% when combined with ArcFace, and 95.78\% TAR at 0\% FAR with CosFace, while providing at least 115-bit security against known attacks.
\end{abstract}

\section{Introduction}\label{sec:intro}
Biometric-based authentication has long been the subject of intensive research, and deep convolutional neural networks (CNNs) have allowed significant recent advances in recognition. The success of the CNN architecture comes from its ability to extract discriminative features from user biometric traits, such as facial images. Within face recognition systems, user face images are well-compressed into deep face templates. Therefore, the exposure of user face templates constitutes a severe threat to user privacy as well as to system security. For instance, face images can be reconstructed from the corresponding templates~\cite{MCY+18} and this is a direct threat to user privacy. In addition, the reconstructed face image can be used to gain unauthorized access to the system. A high number of attacks have already been reported, which shows how risky the leakage of templates is~\cite{MCY+18,CLM07,GRG+13,FJR15,CJ15}.

The goal of face template protection is to satisfy \emph{irreversibility}, \emph{revocability}, and \emph{unlinkability} properties without significant degradation in face recognition performance. In terms of security, in the event a protected face template is leaked to an adversary, reconstructing the original face template from the compromised template should be computationally infeasible. Thus, the compromised template must be revoked and generation of a new protected face template should be feasible. In addition, two or more protected templates created from the same user biometric traits should hide relations between them. Furthermore, the template protection scheme is expected to maintain high matching performance with high intra-user compactness and inter-user discrepancy. Achieving both high template security and high match performance at the same time is a fundamental challenge in the design of face template protection schemes.
\begin{figure*}
\begin{center}
\includegraphics[width=17.4cm, height=6.3363cm]{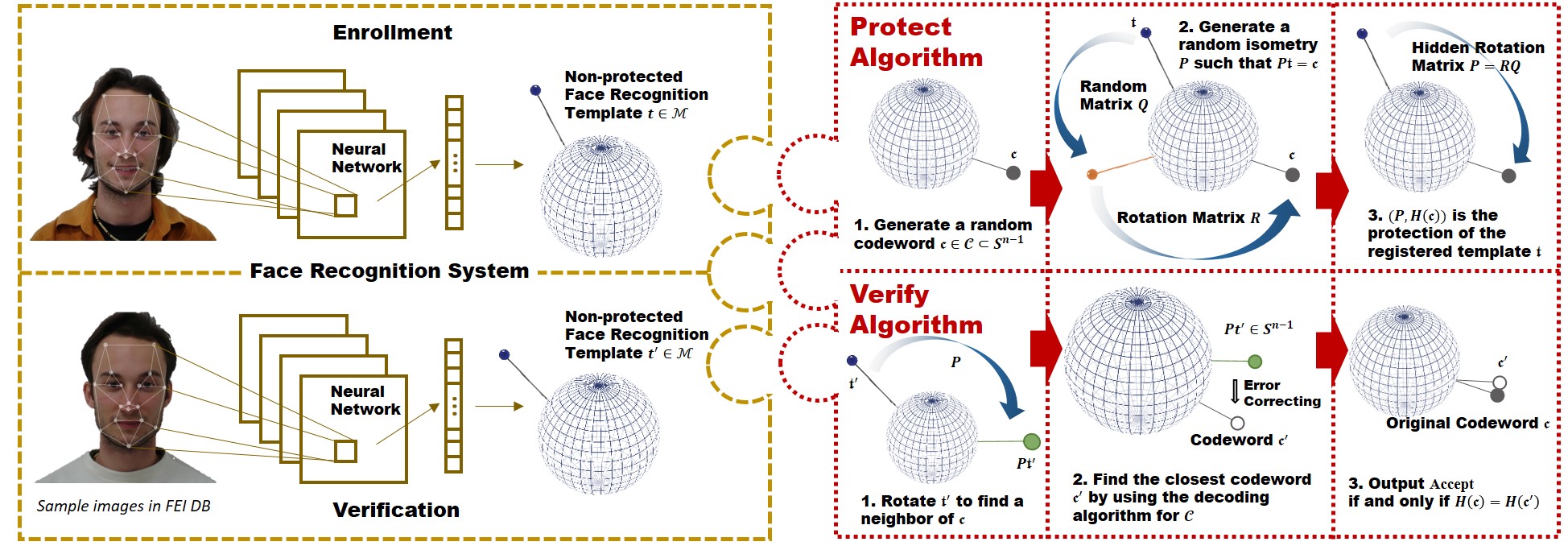}
\end{center}
\vspace{-0.3cm}
   \caption{\small{IronMask is a template protecting modular architecture, which can be combined with any face recognition systems using angular distance metric.}}
\label{fig:intro_ficture}
\vspace{-0.1cm}
\end{figure*}

Most of recent research on CNN-based face template protection can be categorized into two groups, \emph{training-in-enrollment} and \emph{training-before-enrollment}. The former approach~\cite{PZK+16,JCJ18,JCJ19} requires face images of the target user to train the CNN, which maps from face images into binary code during enrollment; at this stage, the resulting binary codes are randomly generated and so have no correlation with the original face images. The hashed value of the resulting binary code is stored as a protected template of the target user. This approach is incompatible with dynamic enrollment systems because, once the initial enrollment has been completed, the CNN would need to be retrained from scratch whenever a new user is added to the system. The latter approach~\cite{TVN19,MST+19} was proposed to mitigate the effect of the above problem. In this case, the CNN can be pre-trained before enrollment, and performance degradation is minimized using additional methods, e.g., neural network based error-correction~\cite{TVN19} and significant feature extraction~\cite{MST+19}. However, compared to the training-in-enrollment approach, these algorithms suffer from low matching accuracy. The common feature of both approaches is the use of a binary code either as an intermediate step or as the final protected template. This artificial binarization makes template-protected face recognition systems quite different from state-of-the-art (non-protected) face recognition systems in terms of both architecture and/or matching performances.

\subsection{Our Contributions}

We present a novel design architecture that maintains the advantages of state-of-the-art (non-protected) biometric recognition as much as possible, while adding high template security. To this end, we propose a modular architecture, called IronMask, that provides template protection to any (non-protected) face recognition system using angular distance metric. A pictorial explanation of the approach is provided in Fig.~\ref{fig:intro_ficture}. By combining IronMask with well-known face recognition systems, we achieve the best matching performance in the training-before-enrollment category. Our performance results are even comparable with existing results in the training-in-enrollment category.

In this paper, we consider an arbitrary biometric recognition system, which measures the closeness of two templates by angular distance metric (or cosine similarity). Two biometric templates $\tp_1,\tp_2$ are considered as vectors on the unit $n$-sphere $S^{n-1}$, and the distance between them is defined as the angle $\cos^{-1}(\langle\tp_1,\tp_2\rangle)$ between them, where $\langle\tp_1,\tp_2\rangle$ denotes the standard inner product of $\tp_1$ and $\tp_2$. It is worth noting that the angular distance~\cite{GY14,LWZ+16,LLW17,RCC17,WXJ+17,LWY+17,FWL+18,WWZ+18,DGX+19,ZZQ+19} has been widely used along with the Euclidean distance in many recent successful face recognition systems to measure closeness between two templates.

In developing IronMask, we devise a new real-valued error-correcting-code (ECC) that is compatible with the angular distance metric and enables to use real-valued codes instead of binary codes, unlike prior works. Hence, given an original template $\tp\in S^{n-1}$, IronMask generates a real-valued codeword~$\tc$ and a linear transformation $\helper$ mapping from $\tp$ to $\tc$, which has the form of an $n$-by-$n$ matrix. The hashed value $\hash(\tc)$ of the codeword and $\helper$ are stored as the protected template, where $\hash$ is a collision resistant hash function. If we set $\helper$ to be an orthogonal matrix, then it becomes an \emph{isometry} that preserves the inner product, so that any noisy template with distance $d$ from $\tp$ can be transformed via $\helper$ into the one with the same distance from $\tc$. Thus, no additional performance degradation is imposed by this transformation. Finally, with the help of the error-correcting capability of our ECC, we can recover the original codeword and check its originality with $\hash(\tc)$. 

If the protected template is leaked, the adversary gains only $\helper$ and $\hash(\tc)$. We inject sufficient randomness into $\helper$ to hide information on the original template via the mathematical tool of choosing a uniform orthogonal matrix. Then, we show that the protected template satisfies the three required properties, irreversibility, revocability, and unlinkability under appropriate parameter sizes.

IronMask can be combined with arbitrary face recognition systems using angular distance metrics. To demonstrate the efficacy of IronMask, we apply IronMask to two well-known face recognition systems, ArcFace~\cite{DGX+19} and CosFace~\cite{WWZ+18}, with three widely used datasets, CMU Multi-PIE~\cite{SBB12}, FEI~\cite{TG10}, and Color-FERET~\cite{PWH+98}. Here, ArcFace and CosFace are one of the state-of-the-art face recognition systems.
Our experiments show that IronMask with ArcFace achieves 98.76\% true accept rate~(TAR) at 0\% false accept rate~(FAR), and with CosFace achieves 81.13\% TAR at 0\% FAR.
Through additional processes involving multiple face images as input of IronMask, we improve the 
matching performance up to 99.79\% TAR at 0.0005\% FAR for ArcFace and 95.78\% TAR at 0\% FAR for CosFace.

\subsection{Related Works}

Many biometric template protection techniques have been proposed in the literature.
Secure sketch~\cite{DRS04}, fuzzy commitment~\cite{JW99}, and fuzzy vault~\cite{JS06} are biometric cryptographic schemes that not only enable rigorous analysis on template security but also guarantee high template security. These cryptographic solutions employ ECCs and cryptographic hash functions as main building blocks. In the cryptographic approach, an original template is mapped onto a random codeword, and its hash value is subsequently stored as the protected template. These schemes vary according to the metric of the space containing the templates. Many face template protection proposals~\cite{SLM07,AL09,WQ10,TVN19,MST+19} follow this approach. However, this requires to use binary codes or points, which leads to loss of discriminatory information on the original template and degradation of matching performance. Biohashing~\cite{TGN06}, cancelable biometric~\cite{RCC+07}, and robust hashing~\cite{TFM+07} are feature transformation approaches that transform the original template into an element in a new domain using a non-invertible transformation and salting. A hybrid approach for face template protection that combines the cryptographic approach with the feature transformation approach was proposed in~\cite{FYJ09}. Local region based hashing was suggested as a way to extract discriminative features from face images~\cite{PG15}; however, the feature extraction consists of quantization and cryptographic hashing, so that this algorithm still suffers from low matching accuracy in the same way as the cryptographic approach.
To improve matching performance as well as security, several studies have proposed CNN-based approaches~\cite{PZK+16,JCJ18,JCJ19}, which minimize intra-user variability and maximize inter-user variability using neural networks. These works~\cite{PZK+16,JCJ18,JCJ19} essentially require one or multiple face image(s) of the target user when training the neural networks, and this requirement is impractical in some applications such as dynamic enrollment systems.

There is another approach~\cite{B18,EJB20} based on fully homomorphic encryption that provides a strong security against attacks using biometric distribution. However, unlike aforementioned works and ours, its security is guaranteed only under the additional assumption that the decryption key of fully homomorphic encryption is kept secret.




\section{Modular Architecture for Protecting Templates}\label{sec:scheme}

The more the technology of face recognition advances, the greater the number of discriminative features that are well-compressed into a face template. Ironically, this phenomenon can be demonstrated by applying previously proposed face image reconstruction attacks to the state-of-the-art face recognition system. For example, we apply the face image reconstruction attack that was originally proposed for attacking Facenet~\cite{SKP15}, to ArcFace~\cite{DGX+19},
which is one of the state-of-the-art face recognition system. 
It shows that the success probability of the attack against ArcFace is higher than that against FaceNet.
Refer to Appendix~A for the details.


In this section, we propose a modular architecture for protecting angular distance based templates including those of ArcFace. 
More precisely, we present a transformation, called IronMask, which transforms a (non-protected) biometric recognition system using angular distance metric into a template-protected one. Biometric recognition systems can be classified into `identification' and `verification' according to usage scenarios. The goal of the identification is to identify a given template among templates in the repository. On the other hand, the verification is equivalent to the authentication. For the sake of simplicity, in this paper we focus on the verification, but note that the proposed protection method also works for the identification. IronMask is a transformation of the verification system and consists of two algorithms, template protection algorithm $\PT$ and verification algorithm $\VF$. The $\PT$ algorithm takes an original template as an input and returns a protected template. Given a template~$\tp$ and a protected template~$\ptp$, intuitively, it is non-trivial to check the closeness between $\tp$ and $\ptp$, so that a specialized verification algorithm~$\VF$ is necessary in our approach. The $\VF$ algorithm takes a non-protected template~$\tp'$ and a protected template $\ptp$ where $\ptp$ is the output of the $\PT$ algorithm with input $\tp$, and then it outputs $1$~(Accept) or $0$~(Reject) according to the closeness between $\tp$ and $\tp'$.

\vspace{-0.1cm}
\paragraph{Mathematical Concepts.} We introduce some mathematical concepts used in the paper. A cryptographic hash function, denoted by $\hash$, is a one-way function such as SHA-256 that given a hash value, it is computationally infeasible to find its preimage. An error correcting code (ECC) is a set of codewords, $\mathcal{C}$, in the metric space~$\M$ with decoding function~$\mathsf{Decode}$ such that, informally speaking, given an element in $\M$, it finds the closest codeword in $\mathcal{C}$. In this paper, we are interested in the case $\M=S^{n-1}$, the set of all unit vectors in $\mathbb{R}^n$. An orthogonal matrix is a square matrix whose columns and rows are orthonormal vectors. 
%
Throughout the paper, $O(n)$ denotes the set of $n\times n$ orthogonal matrices.

\vspace{-0.1cm}

\paragraph{IronMask: Transformation for Template Protection.} We observe that templates in our target recognition systems are vectors in the unit $n$-sphere $S^{n-1}$ with angular distance. 
Here, our basic idea for transformation is to use an orthogonal matrix, which is an isometry keeping angular distance between templates once transformed. Then, the orthogonal matrix becomes a part of protected template. To generate this special orthogonal matrix, we define an algorithm $\mathsf{HRMG}$ that takes two inputs $\mathbf{a},\mathbf{b}\in S^{n-1}$ and returns a random orthogonal matrix $\helper$ such that $\mathbf{b}=\helper\mathbf{a}$. This algorithm will be specified in Section~\ref{sec:distribution}. In addition, we use a sub-algorithm, called $\mathsf{USample}$, to describe an abstract construction of IronMask. The $\mathsf{USample}$ algorithm generates a uniform element from the set of codewords and the details will be described in Section~\ref{sec:ECC}. Now we present the abstract construction of IronMask in Algorithms~\ref{alg:protect} and~\ref{alg:verify}. 

\vspace{-0.2cm}
\begin{algorithm}
\SetAlgoLined
\SetKwInOut{Input}{Input}
\SetKwInOut{Output}{Output}
\Input{Template $\tp\in\M$}
\Output{Protected template $\ptp$}
\nl $\tc\leftarrow\mathsf{USample}$

\nl $\helper\leftarrow\mathsf{HRMG}(\tp,\tc)$\ such\ that\ $\tc = \helper\tp$\ and $\mathrm{r}\leftarrow\hash(\tc)$.

\nl Output $\ptp:=(\mathrm{r},\helper)$.
 \caption{Protect Algorithm}\label{alg:protect}
\end{algorithm}

\vspace{-0.8cm}

\begin{algorithm}
\SetAlgoLined
\SetKwInOut{Input}{Input}
 \SetKwInOut{Output}{Output}
\Input{Protected template $\ptp$ and template $\tp'\in\M$}
\Output{$Accept$ or $Reject$}
\nl Parse $\ptp$ as $(\mathrm{r},\helper)$.

\nl $\tc'\leftarrow\mathsf{Decode}(\helper\cdot \tp')$ and $ \mathrm{r'}\leftarrow\hash(\tc')$.

\nl Output $Accept$ if $\mathrm{r' =r}$. Otherwise, output $Reject$.
 \caption{Verify Algorithm}\label{alg:verify}
\end{algorithm}
\vspace{-0.2cm}

By the definition of $\mathsf{HRMG}(\tp,\tc)$, $\helper$ is an isometry transformation from the original template $\tp$ to the codeword $\tc$. That is, since the orthogonal matrix $\helper$ preserves inner products, the angular distance between the original template $\tp$ and a newly recognized template $\tp'$ is equal to that between $\tc = \helper\tp$ and $\tc' = \helper\tp'$. Therefore, if both $\tp$ and $\tp'$ are created from the same user, then we expect that $\tc$ and $\tc'$ have a small angular distance, so that $\mathsf{Decode}(\tc')$ returns $\tc$ with high probability where $\mathsf{Decode}$ is the decoding function of the employed ECC. 

Our abstract construction of IronMask is quite simple, but its realization is challenging mainly due to the following reasons:
\begin{enumerate}\itemsep1mm
    \item We need a suitable candidate for ECC over $S^{n-1}$ with angular distance.
    \vspace{-0.2cm}
    \item The $\mathsf{HRMG}$ algorithm should prevent from leaking an important information on the input template~$\tp$. 
    \vspace{-0.1cm}
\end{enumerate}

In the next subsections, we will resolve the above two issues to complete our IronMask construction.


\subsection{Error Correcting Codes over $S^{n-1}$ with Angular Distance}\label{sec:ECC}
In this subsection, we design a new real-valued ECC, which is a special building block for IronMask and quite different from usual binary ECCs such as BCH codes. To clarify the design rationale, we first list requirements that ECCs for IronMask should satisfy.

\vspace{-0.1cm}
\begin{itemize}
\item (Discriminative) All codewords are well-spread on $S^{n-1}$, so that any two of codewords are sufficiently far to each other. This property is necessary for high accuracy of IronMask.
\vspace{-0.3cm}
\item (Uniformly-Samplable) There exists an efficient algorithm $\mathsf{USample}$ generating a codeword with uniform distribution over the set of all codewords. This property will be used in the $\PT$ algorithm of IronMask. 
\vspace{-0.6cm}
\item (Efficiently-Decodable) There exists an efficient algorithm $\mathsf{Decode}$ that takes any vector in $S^{n-1}$ as an input and returns a codeword with the shortest angular distance from the input vector. This property will be used in the $\VF$ algorithm of IronMask. 
\vspace{-0.2cm}
\item (Template-Protecting) The number of codewords is sufficiently large in order to prevent the brute-force attack, e.g., $2^{80}$ for $80$ bits security.
\end{itemize}

Next, we devise a new ECC satisfying all the above four requirements. To this end, we first define a family of codeword sets $\{\mathcal{C}_\th\}_\th$.

\vspace{0.2cm}
\noindent \textbf{Definition 1.} \textit{For any positive integer $\th$, $\mathcal{C}_\th$ is defined as a set of all unit vectors whose entries consist of only three real numbers $-\frac{1}{\sqrt{\th}}$, $0$, and $\frac{1}{\sqrt{\th}}$. 
Then, each codeword in $\mathcal{C}_\th$ has exactly $\th$ nonzero entries.}

\vspace{0.2cm}
In order to handle codewords for our purpose, we develop two algorithms, called $\mathsf{Decode}$ and $\mathsf{USample}$. The $\mathsf{Decode}$ algorithm takes an arbitrary unit vector in $S^{n-1}$ as an input and returns a codeword that is the closest to the input vector. The $\mathsf{USample}$ algorithm efficiently samples a codeword with uniform distribution over $\mathcal{C}_\th$. The detailed descriptions of both algorithms are presented in Algorithms~\ref{alg:decode} and~\ref{alg:usample}, respectively.

We present a useful proposition showing features of the proposed ECC $\mathcal{C}_\th$ in terms of our requirements.\\


\noindent \textbf{Proposition 1.} \textit{Let $\mathcal{C}_\th$ be the set of codewords defined in Definition 1. 
\begin{enumerate}
    \item The minimum angular distance between any two distinct codewords in $\mathcal{C}_\th$ is $\cos^{-1}(1-\frac{1}{\th})$.
    \vspace{-0.2cm}
    \item The output of the $\mathsf{USample}$ algorithm is uniformly distributed over $\mathcal{C}_\th$.
    \vspace{-0.2cm}
    \item For any $\tu\in S^{n-1}$, the output $\tc$ of the $\mathsf{Decode}$ algorithm satisfies the following inequality:
$$\langle \tc,\tu\rangle\geq\langle \tc',\tu\rangle\text{ for all }\tc'\in\mathcal{C}_\th.$$
    \vspace{-0.3cm}
    \item The number of all codewords in $\mathcal{C}_\th$ is $\binom{n}{\th}\cdot2^{\th}$. 
\end{enumerate}}
%
%

\begin{algorithm}
\SetAlgoLined
\SetKwInOut{Input}{Input}
\SetKwInOut{Output}{Output}
\Input{$\tu=(u_1,\ldots,u_n)\in S^{n-1}$}
\Output{$\tc=(c_1,\ldots,c_n)\in S^{n-1}$ such that $\langle \tc,\tu\rangle\geq\langle \tc',\tu\rangle\text{ for all }\tc'\in\mathcal{C}_\th$}
\nl Find a set of indices $J\subset[1,n]$ of size $\th$ such that $\forall j\in J$ and $\forall k\in[1,n]\setminus J$, $|u_j|\geq|u_k|$.

\nl $\textbf{if}\ j\in J\ \textbf{then}\ c_{j}\leftarrow \frac{u_{j}}{|u_{j}|\sqrt{\th}}$\\
$\textbf{else}\ c_{j}\leftarrow 0$

\nl Output $\tc:=(c_1, \ldots, c_n)$.
 \caption{$\mathsf{Decode}$ Algorithm}\label{alg:decode}
\end{algorithm}
\begin{algorithm}
\SetAlgoLined
\SetKwInOut{Input}{Input}
\SetKwInOut{Output}{Output}
\Input{Randomness seed}
\Output{$\tc=(c_1,\ldots,c_n)\in S^{n-1}$ that is uniformly sampled from $ \mathcal{C}_\th$}
\nl Choose $\th$ distinct positions $j_1,\ldots,j_{\th}$ from $\{1,\ldots,n\}$ and set $J:=\{j_1,\ldots,j_{\th}\}$.

\nl For $j\in J$, set $c_j$ to one of $-\frac{1}{\sqrt{\th}}$ and $\frac{1}{\sqrt{\th}}$ at random.\\
For other indices $j\not\in J$ set $c_j:=0$.

\nl Output $\tc:=(c_1,\ldots,c_n)$.
 \caption{$\mathsf{USample}$ Algorithm}\label{alg:usample}
\end{algorithm}

%
Due to the space constraint, the proof of Proposition 1 is relegated to Appendix~B.

The above proposition tells us that the proposed ECC, $\mathcal{C}_\th$, is a suitable candidate for ECCs in IronMask.
The first result gives the minimum angular distance of $\mathcal{C}_\th$ and it 
confirms that $\mathcal{C}_\th$ is fairly \emph{discriminative}.
However, it is still unclear, as of yet, whether such the minimum distance is sufficient for real world applications. To clarify it, we will complement this argument both experimentally~(in Section~\ref{sec:exp}) and theoretically~in Appendix~B.
The second and third results show that $\mathcal{C}_\th$ is \emph{uniformly-sampleable} and \emph{efficiently-decodable}, respectively.
In practical face recognition systems such as ArcFace~\cite{DGX+19} and CosFace~\cite{WWZ+18},
$n$ is set to be $512$. Thus, if $\th$ is set appropriately, the fourth result shows that $\mathcal{C}_\th$ satisfies the last requirement. For example, the number of all codewords in $\mathcal{C}_{16}$ with $n=512$ is larger than $2^{115}$.

\subsection{Hidden Rotation Matrix Generation}\label{sec:distribution}
In this subsection, we construct an $\mathsf{HRMG}$ algorithm that takes two input vectors~$\tp$, $\tc$ in $S^{n-1}$ and returns an isometry which preserves angular distance (equivalently, inner product) as well as maps from $\tp$ to $\tc$.

\paragraph{Naive Approach.}
First, we present a naive approach to generate such an isometry, a rotation in the plane uniquely defined by two input vectors $\tp$ and $\tc$. More precisely, let $\mathbf{w}=\tc-\tp^{T}\tc\tp$ and $\mathbf{R}_{\theta}=\left[\begin{array}{rr} \cos{\theta} & ~-\sin{\theta} \\ \sin{\theta} & ~\cos{\theta} \end{array}\right]$ where $ \theta=\cos^{-1} (\langle\tp, \tc\rangle)$. Then, the naive isometry $\mathbf{R}$ mapping from $\tp$ to $\tc$ can be precisely calculated as follows:
\begin{align*}
\mathbf{R}=\underbrace{\mathbf{I}-\tp\tp^{T}-\mathbf{w}\mathbf{w}^{T}}_{\textrm{projection part}}+\underbrace{[\tp\ \mathbf{w}]\mathbf{R}_{\theta}[\tp\ \mathbf{w}]^{T}}_{\textrm{rotation part}}
\end{align*}
where $\mathbf{I}$ is the $n\times n$ identity matrix.\\
%
%
\\
\noindent \textbf{Proposition 2.} Given $\tp$ and $\tc$, \textit{the above matrix $\mathbf{R}$ is an orthogonal matrix such that $\mathbf{R}\tp=\tc$.}\\

The proof of Proposition 2 is given in Appendix~B.

\begin{figure}[!h]
\begin{center}
    \includegraphics[width=7.0cm, height=4.8cm]{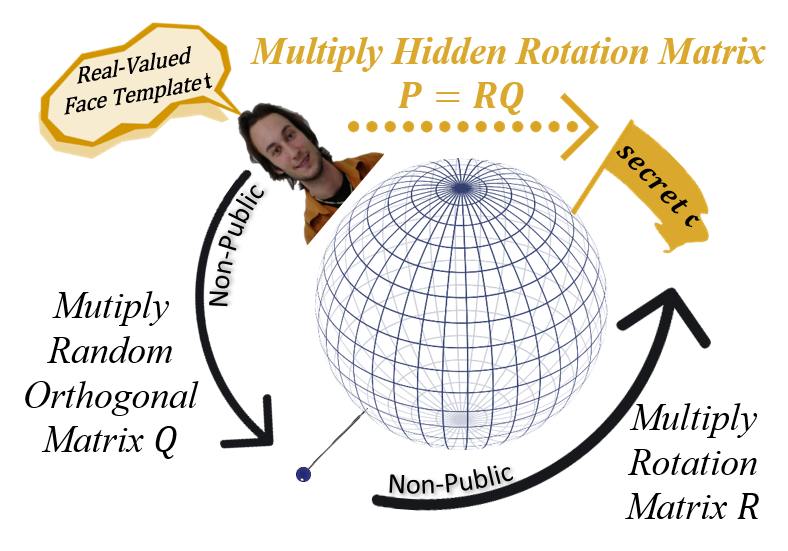}
\end{center}
\vspace{-0.5cm}
\outfigcaption{\small{Geometrical interpretation of an orthogonal matrix~$\mathbf{Q}$, a rotation matrix~$\mathbf{R}$, and the final result~$\mathbf{P}$}}\label{fig:HRMGfigure}
\end{figure}

\paragraph{Template-Protecting Approach.}
Although the above naive algorithm functions correctly, from the privacy point of view it might leak some information on input vectors $\tp$ and $\tc$ because it only rotates on the plane that includes both $\tp$ and $\tc$ which should be kept secret in our application, biometric recognition system. As aforementioned, we expect that the algorithm $\mathsf{HRMG}(\tp,\tc)$ hides information on $\tp$ as much as possible. To this end, we add a randomizing step to blind two input vectors while preserving the necessary property for mapping from $\tp$ to $\tc$. The proposed $\mathsf{HRMG}$ algorithm is described in Algorithm~\ref{alg:hrmg} and a pictorial explanation for the $\mathsf{HRMG}$ algorithm is given in Figure~\ref{fig:HRMGfigure}.

\vspace{-0.15cm}

\begin{algorithm}
\SetAlgoLined
\SetKwInOut{Input}{Input}
\SetKwInOut{Output}{Output}
\Input{A pair of template and codeword $(\tp, \tc)$}
\Output{An isometry $\helper\in O(n)$ mapping $\tp\mapsto\tc$}
\nl Choose an orthogonal matrix $\mathbf{Q}\in O(n)$ at random.

\nl Compute a rotation matrix $\mathbf{R}$ mapping $\mathbf{Q}\tp\mapsto\tc$.

\nl Output $\helper := \mathbf{R}\mathbf{Q}$.
 \caption{$\mathsf{HRMG}$ Algorithm}\label{alg:hrmg}
\end{algorithm}

As for the first step of the $\mathsf{HRMG}$ Algorithm, we use a well-known algorithm~\cite{Ste80} which generates a random orthogonal matrix~$\mathbf{Q}$ from the uniform distribution. As for the second step to generate $\mathbf{R}$, we use a naive rotation matrix generation given at the beginning of this subsection. 


\section{Security Analysis}
In this section, we look into the security of IronMask, proposed in Section~\ref{sec:scheme}.
For this purpose, we first introduce security requirements for biometric recognition system, and then present plausible attack scenarios and their countermeasures. 


\subsection{Security Requirements}
For the security of biometric recognition systems, the following three conditions are required to be satisfied.
\vspace{-0.2cm}
\begin{itemize}
%
	\item Irreversibility: It is computationally infeasible to recover the original biometric data from the protected template.
	\vspace{-0.2cm}
	\item Revocability: It is possible to issue new protected templates to replace the compromised one.
	\vspace{-0.2cm}
	\item Unlinkability: It is computationally infeasible to retrieve any information from two protected templates generated in two different applications.
\end{itemize}

In IronMask, the $\PT$ algorithm outputs a pair of a hash output of a randomly selected codeword and an orthogonal matrix which maps from a biometric template to the selected codeword. So, we assume that the adversary can access to a database that stores those pairs, which were generated by the $\PT$ algorithm. 

\subsection{Plausible Attacks and Their Countermeasures}
Now, we look into irreversibility, revocability, and unlinkability of IronMask by describing several plausible attacks and then presenting countermeasures. 

\paragraph{Irreversibility.} 
Given a pair of $(\mathrm{r}, \helper)$, the goal of the adversary who wants to break the irreversibility of IronMask, is to recover the biometric template $\tp$, which was the input of the $\PT$ algorithm once the given $(\mathrm{r}, \helper)$ was generated. Then, they hold that $\tc = \helper\tp$ and $\mathrm{r} = \hash(\tc)$ for the exploited hash function $\hash$. Since $\helper$ is an invertible matrix,
there exists a unique $\tp$ for each $\tc$ in the set of codewords, $\mathcal{C}_\th$. However, if the distribution of $\tp$ is hidden, the only way to check whether $(\tp, \tc)$ is the solution to $\tc = \helper\tp$ and $\mathrm{r} = \hash(\tc)$, is to check whether $\mathrm{r}$ is $\hash(\tc)$ or not. 
On the other hand, there is no known result about the distribution of biometric templates generated by the face recognition systems that we consider in this paper. So, it is reasonable to assume that the distribution of $\tp$ is hidden to the adversary.
Therefore, under assuming that the distribution of $\tp$ is hidden, the exploited hash function $\hash$ is one-way, and the cardinality of $\mathcal{C}_\th$ is sufficiently large, e.g., larger than $2^{80}$, it is computationally infeasible to recover $\tp$ from the given $(\mathrm{r}, \helper)$.

\paragraph{Revocability.}
For the revocability, it should be guaranteed that the irreversibility holds with a newly issued protected template, even though the old protected templates generated by the same biometric source are given to the adversary.
In our setting, since the underlying face recognition systems are based on CNN, once a user would like to generate a new protected template again for the revocation, the unprotected template $\tp$ is newly generated. It is different from the previously generated ones, but the distances between them are very close. By considering this feature, the revocability can be regarded that any $\tp_i$'s should not be revealed
once several $(\mathrm{r}_i, \helper_i)$'s are given, where $(\mathrm{r}_i, \helper_i)$ is the outcome of 
the $\PT$ algorithm with input $\tp_i$ and each $\tp_i$ is generated from the same biometric source.

Because IronMask satisfies the irreversibility, it is computationally infeasible to recover $\tp_i$ from given a pair of $(\mathrm{r}_i, \helper_i)$ only.
However, by using several pairs of $(\mathrm{r}_i, \helper_i)$, the adversary may obtain relations $\helper_i^{T}\tc_i = \helper_j^{T}\tc_j+\te_{ij}$ for each $i, j$ where $\te_{ij} = \tp_i-\tp_j$ since the relations $\tc_i = \helper_i\tp_i$, $\tc_j = \helper_j\tp_j$ hold and $\helper_i$'s are orthogonal matrices.
It is equivalent to $\tc_i = \helper_i\helper_j^{T}\tc_j+\helper_i\te_{ij}$ and thus we replace breaking the revocability by finding $(\tc_i, \tc_j, \te_{ij})$ such that 
\begin{eqnarray}\label{eqn:revocability}
\helper\tc_j+\helper_{i}\te_{ij}=\tc_i~\text{with}~\te_{ij}= \tp_i - \tp_j,\nonumber \\ 
~\helper_i\tp_i = \tc_i\in\mathcal{C}_\th, \ ~\text{and}~\helper_j\tp_j = \tc_j\in\mathcal{C}_\th
\end{eqnarray}
for given orthogonal matrices $\helper_i$, $\helper_j$, and $\helper=\helper_i\helper_j^{T}$. Here, we again remark that $\tc_i, \tc_j$ are codewords and so they are sparse unit vectors of a fixed small weight whose non-zero elements have the same absolute value.


A naive way to resolve the above problem is to check all possible candidates, but it is computationally infeasible if the cardinality of the set of codewords is exponentially large in the security parameter. Someone who is familiar with mathematical structures may consider to apply structural attacks, such as lattice reductions. However, unfortunately, since $\helper_i$'s (and thus $\helper$) are orthogonal matrices, a lattice reduction with input $\helper_i$ returns $\helper_i$ itself.

Also, someone who is familiar with cryptanalysis may consider to apply a time-memory-trade-off~(TMTO) strategy whose complexity is approximately a square-root of the cardinality of the search space.
However, $\te_{ij}$ is also unknown and its space is very huge since it is a vector of small real numbers. Thus, the search space for $(\tc_i, \tc_j, \te_{ij})$ is sufficiently large under appropriate parameters and this type of attacks does not work well. We elaborate the details of the TMTO strategy to solve Equation~\eqref{eqn:revocability} in Appendix~C.



\paragraph{Unlinkability.}
For the unlinkability, the adversary should not find any information of the biometric data from two protected templates generated in two different applications.
Suppose that two protected templates $(\mathrm{r}_1, \helper_1)$ and $(\mathrm{r}_2, \helper_2)$ are given to the adversary.
If they are generated from the same biometric source, it has a solution to Equation~\eqref{eqn:revocability}, as in the case of revocability.
That is, the adversary knows that there is a solution to Equation~\eqref{eqn:revocability}
if he can break the unlinkability of IronMask.
To the best of our knowledge, however, the best way to decide the existence of such a solution to Equation~\eqref{eqn:revocability}, is to find it directly. It is infeasible under our parameter setting, as in the case of revocability. Thus, IronMask satisfies the unlinkability.



\definecolor{aqua}{rgb}{0.0, 1.0, 1.0} 
\definecolor{brilliantlavender}{rgb}{0.96, 0.73, 1.0} 

\section{Performance Evaluation}\label{sec:exp}
In this section, we investigate implementation aspects of IronMask combining with main CNN-based face recognition systems.

\subsection{Face Datasets and Recognition Algorithm}\label{subsec:Face}
We use three popular datasets targeted face recognition to demonstrate the feasibility of IronMask. Below we briefly explain which face images were used in each dataset for our experiments. 
\vspace{-0.1cm}
\begin{itemize}
\item CMU Multi-PIE Dataset~\cite{SBB12}: It consists of 750,000 face images of 337 subjects. The photos are captured under 4 different sessions, 15 different rotated view points, and 19 illumination conditions. We selected 20 face images randomly in the sessions 3 and 4 for 194 common subjects each. Then, we used 5 poses (p04, p05.0, p05.1, p13, p14) and 2 illumination conditions (i08, i18) for each session, respectively. 
\vspace{-0.3cm}
\item FEI Dataset~\cite{TG10}: It consists of 2,800 face images of 200 subjects. The photos are captured under 14 different rotated view points for each subject. We selected 11 poses (p02, p03, p04, p05, p06, p07, p08, p09, p11, p12, p13) among them for each subject in our experiments. 
\vspace{-0.3cm}
\item Color-FERET Dataset~\cite{PWH+98}: We used 4 face images for 237 subjects each where the photos are varied with respect to pose, illumination, expression, and glasses.
\end{itemize}
\vspace{-0.1cm}

In order to show compatibility of our proposed modular architecture, we implement it over two notable face recognition systems using angular distance metric. One of them is CosFace in~\cite{WWZ+18} and the other is ArcFace in~\cite{DGX+19}. We note that ArcFace that achieves the state-of-the-art performance on the NIST Face Recognition Vendor Test~(FRVT)\footnote{\url{https://github.com/deepinsight/insightface}}.
In our experiments, we use ResNet50 with 512-dimensional output as a backbone scheme for both ArcFace and CosFace.

In many real world applications, multiple face images for a single user are available during the enrollment, so that multiple templates are also available once applying face recognition. As an intermediate process before applying our IronMask, we find a kind of center point of multiple face templates and use it as the final template of user, where the center point has a minimal cost in inner products with multiple templates.\footnote{Each template is considered as a vector in $S^{n-1}$ with cosine similarity distance. Therefore, simple averaging with the Euclidean distance cannot generate a meaningful center and we need to devise a different center-finding algorithm.} This procedure contributes to both 
high compactness for intra-class and small degradation of discrimination for inter-class.
We experimentally show that IronMask is well compatible with not only the original face recognition systems but also the above intermediate process with multiple images.
Multiple linear regression (MLR) is sufficient for realizing the above intermediate process to find the center because the center is linearly defined from multiple templates. In our implementation, MLR is fine-tuned, where weights are normally initialized and the cost function with L1 loss is optimized using stochastic gradient descent~(SGD) with learning rate (0.01, 0.001) during (1000, 500) epochs. 
We normalize the output of MLR and then use it as the final template, which is an input of the $\PT$ algorithm.

\subsection{Experimental Results}\label{sec:Eval}

We provide our experimental results of IronMask over ArcFace and CosFace with the aforementioned three datasets. We consider the single-image setting and the multi-image setting, where the latter applies the intermediate process that uses a center point and the former does not.
For the single-image setting, we use two standard metrics, a true accept rate~(TAR) and a false accept rate~(FAR), to evaluate matching performance of IronMask. For the multi-image setting, only a limited number of face images per each user is available in the target databases. Thus, in order to reasonably evaluate matching performance of IronMask, we carefully design experiments as follows: 1) We randomly choose $n_1$ face images among $n_0$ total face images for each user to extract a centered input feature of the $\PT$ algorithm. 2)  We run the $\VF$ algorithm with $n_2~(= n_0-n_1)$ face images of the same user for TAR and $n_3~(= n_0\times (k-1)) $ face images of different users for FAR, where $k$ is the number of all users in each dataset. 3) We run Steps 1) and 2) $r$ times repeatedly to get more reliable performance results. That is, we have $n_2 \times r$ TAR and $n_3 \times r$ FAR tests.
For the parameter $(k,n_0,n_1,n_2,n_3,r)$, we choose $(194,20,5,15,20\times193,10)$ for the CMU Multi-PIE dataset, $(200,11,5,6,11\times 199,10)$ for the FEI dataset, and $(237, 4, 2, 2$, $4\times 238, 6)$ for the Color-FERET dataset according to Section~\ref{subsec:Face}.

The graphs of Figure~\ref{fig:Plots} present our experimental results and Table~\ref{table:Performance} summarizes TARs/FARs with respect to face recognition system and dataset.  
Our construction with ArcFace (resp. CosFace) achieves 99.79\% (resp. 95.78\%) TAR at 0.0005\% (resp. 0\%) FAR when using the proposed ECC $\mathcal{C}_{16}$ with $n=512$ (resp. $n=512$) that provides at least 115 bits of security level. We further improve the matching performance by using centered templates as inputs of the $\mathsf{Protect}$ algorithm, as described in Section~\ref{subsec:Face}. For instance, it provides about 1\% (resp. 22\%) improvement in TAR  at the cost of $0.0013$\% (resp. 0\%) degradation in FAR for the CMU Multi-PIE dataset with ArcFace (resp. CosFace).

\begin{table}[]
\small
\begin{center}
\begin{tabular}{|c||l|l|l|}
\hline
\multicolumn{1}{|c||}{$\textsf{Recog}$} & \multicolumn{1}{c|}{CMU Multi-PIE} & \multicolumn{1}{c|}{FEI} & \multicolumn{1}{c|}{Color-FERET}  \\ \hline
AF          & 99.05@0       & 99.63@0      & 98.13@0      \\ \hline
AF+I        & 98.95@0       & 99.27@0      & 98.06@0      \\ \hline
AF+C        & 99.57@9e-4  & 99.82@0      & 99.78@0      \\ \hline
AF+I+C      & 99.96@1.3e-3  & 99.96@3e-4 & 99.46@0      \\ \hline
CF          & 96.16@6e-4  & 98.69@0      & 96.55@0      \\ \hline
CF+I        & 75.56@0       & 87.52@0      & 80.32@0      \\ \hline
CF+C        & 99.92@1.3e-3  & 99.92@0      & 98.96@2e-4 \\ \hline
CF+I+C      & 97.37@0       & 98.53@0      & 91.45@0      \\ \hline
\end{tabular}
\end{center}
\vspace{-0.4cm}
\caption{\small{TAR@FAR performance of $8$ face recognitions for three databases are given. `AF' and `CF' indicate ArcFace and CosFace, respectively, and `+C' indicates whether applying the intermediate process using multiple face images and `+I' indicates whether applying the proposed template protection method, IronMask.}} \label{table:Performance}
\vspace{-0.3cm}
\end{table}

\input{plots_2.tex}

\begin{table}[!t]
\small
\begin{center}
\begin{tabular}{|c|c|c|l|l|}
\hline
Dataset                      & $\textsf{Alg}$                             & $\textsf{ET}$            & \multicolumn{1}{c|}{$\textsf{OT}$} & \multicolumn{1}{c|}{TAR@FAR} \\ \hline
\multirow{4}{*}{CMU Multi-PIE}   & \multirow{2}{*}{\cite{TVN19}}                  & \multirow{2}{*}{Z} & B255                    & 81.40@1e-2                   \\ \cline{4-5} 
                             &                                            &                    & B1023                   & 81.20@1e-2                   \\ \cline{2-5} 
                             & \multirow{2}{*}{Ours}                      & Z                  & R512                    & 98.95@0                      \\ \cline{3-5} 
                             &                                            & Z(C)               & R512                    & 99.96@1.3e-3                  \\ \hline
\multirow{6}{*}{FEI}         & \multirow{4}{*}{\cite{JCJ19}}                  & \multirow{2}{*}{O} & B256                    & 99.73@0                      \\ \cline{4-5} 
                             &                                            &                    & B1024                   & 99.85@0                      \\ \cline{3-5} 
                             &                                            & \multirow{2}{*}{M} & B256                    & 99.84@0                      \\ \cline{4-5} 
                             &                                            &                    & B1024                   & 99.98@0                      \\ \cline{2-5} 
                             & \multirow{2}{*}{Ours}                      & Z                  & R512                    & 99.27@0                      \\ \cline{3-5} 
                             &                                            & Z(C)               & R512                    & 99.96@3e-4                   \\ \hline
\multirow{6}{*}{Color-FERET} & \multirow{4}{*}{\cite{JCJ19}}                  & \multirow{2}{*}{O} & B256                    & 98.31@0                      \\ \cline{4-5} 
                             &                                            &                    & B1024                   & 99.13@0                      \\ \cline{3-5} 
                             &                                            & \multirow{2}{*}{M} & B256                    & 98.69@0                      \\ \cline{4-5} 
                             &                                            &                    & B1024                   & 99.24@0                      \\ \cline{2-5} 
                             & \multicolumn{1}{l|}{\multirow{2}{*}{Ours}} & Z                  & R512                    & 98.06@0                      \\ \cline{3-5} 
                             & \multicolumn{1}{l|}{}                      & Z(C)               & R512                    & 99.46@0                      \\ \hline
\end{tabular}
\end{center}
\vspace{-0.4cm}
\caption{\small{The performance comparison among CNN-based template protected face recognition systems is given. In the \textsf{ET} (enrollment type) column, `Z', `O', and `M' mean that the number of face images required for training during the enrollment are zero, one, and multiple, respectively. That is, both `O' and `M' belong to the training-in-enrollment category, but `Z' belongs to the training-before-enrollment category. `Ours/Z' indicates the combination of ArcFace~(ResNet50) and IronMask.  `Ours/Z(C)' indicates the combination of ArcFace~(ResNet50), the center finding process (C), and IronMask. In the \textsf{OT} (output type) column, `B' means binary and `R' means real.}
}\label{tab:Comparison}
\vspace{-0.4cm}
\end{table}

One may have doubts that the relatively short minimum distance of the proposed ECC~$\mathcal{C}_\th$ causes serious performance degradation of IronMask.
According to Proposition~1, the minimum distance of $\mathcal{C}_{16}$ is $\cos^{-1}(\tfrac{15}{16})$, which is about 20.36$^\circ$, and so the theoretical error capacity is about $10.18^\circ=20.36^\circ/2$.
In case of ArcFace, the threshold which decides the closeness between two templates is about 37$^\circ$ and thus the gap between ArcFace and ours looks quite large. 
However, since there are only a limited number of pairs of codewords whose distance matches the minimum and most pairs of codewords have a longer distance than the minimum in $\mathcal{C}_{16}$, our $\mathsf{Decode}$ algorithm may decode a longer distance than the minimum distance for many cases. 
We confirm the above feature by comparing the performance between the original face recognition systems and IronMask combined with them. 
In Table~\ref{table:Performance}, IronMask with ArcFace (resp. CosFace) degrades a TAR of 0.18~\% (resp. 16.00~\%) in average at the almost same FAR.



We also compare IronMask over ArcFace with the recent deep neural network based template protection methods. Table~\ref{tab:Comparison} presents comparison results of ours with the state-of-the-art researches for each dataset: \cite{TVN19} for CMU Multi-PIE dataset and~\cite{JCJ19} for FEI and Color-FERET datasets.
The main difference between ours and others is that we use a real-valued face feature vector itself without binarization, which may cause significant deterioration in the matching performance.
In Table~\ref{tab:Comparison}, ours improves a TAR of at least 17.7\% than~\cite{TVN19} with the zero-shot enrollment. 
We note that the zero-shot enrollment does not re-train even if a new identity is enrolled in the authentication system.
Table~\ref{tab:Comparison} also shows that ours has comparable matching performance to the schemes that re-train using one or multiple images in the enrollment system.
For the readers who are interested in time and storage required for running IronMask, we report our experimental results in terms of running times and storage below. 
We have implemented our construction on Linux using a single machine with Intel Core i7-6820 HQ running at 2.70~GHz and 16~GB RAM. 
Our experiments take 250~ms for the $\PT$ algorithm at the Linux machine. For the $\VF$ algorithm, it takes 6.72~ms at the Linux machine.
On the other hand, since the protected template~$\ptp=(\mathrm{r},\helper)$ consists of a hash value of codeword and an $n\times n$ real-valued matrix, the size of the protected template under $\mathcal{C}_{16}$ with $n=512$ is less than 1~Mb for face recognition.
It shows that IronMask is sufficient to be deployed in real world applications such as WLogin services using biometric authentication.



{\small

}


\newpage
\appendix

\begin{center}
\Large
\textbf{Appendix}
\end{center}

\section{Template Attack against ArcFace}\label{app:att2arcface}
\setcounter{table}{2}
A template attack is one of the severest attacks against biometric recognition systems~\cite{CLM07,GRG+13,FJR15,CJ15,MCY+18}. In particular, attacks using deep neural network successfully recover an analog of the original facial image from its template of target recognition systems. The neighborly de-convolutional neural network (NbNet) is a deep neural network based template attack that successfully shows vulnerability of a deep face recognition, Facenet in~\cite{MCY+18}. It achieves over 95.2\% TAR at 0.1\% FAR on the LFW dataset. 
The original NbNet was designed for the neural network architecture of Facenet using the triplet-loss function. We adjust and adapt it to the neural network architecture of ArcFace using additive angular margin loss and obtain a superior result than that applied to Facenet in~\cite{MCY+18}. This result implies that templates of ArcFace are more exposed to threats than those of FaceNet.


The NbNet consists of 6 neighborly de-convolution blocks (NbBlock), similar to Densenet, and 1 convolution operation. It uses two different loss functions. One of them is a pixel-wise loss function and the other is the L2-norm of difference between the outputs of feature mapping function. They empirically determined the second activation function of the third layer as the mapping function. 

In order to adapt the NbNet to ArcFace, we reduce the size of one NbBlock and change the perceptual loss to the angular distance between the outputs of ArcFace model. 
Since the Arface model uses a small variation of margin loss for intra-class and a large variation of margin loss for inter-class, ArcFace uses larger threshold than Facenet. Under this setting, we experiment the vulnerability of ArcFace against template attack and obtain the results in Table~\ref{table:NbNet2ArcFace}.
According to our experimental results, at FAR of 0.1\%, TAR of templates generated by the NbNet for ArcFace is about 2.5\% higher than that for Facenet. 
It is evident that the more advanced face recognition may cause severer threats to user privacy.

\begin{table}[h!]
\centering
\begin{tabular}{|c|c|c|c|}
\hline
FAR     & $0.0001\%$ & $0.1\%$   & $1.0\%$   \\ \hline
\hline
Facenet & NA         & $95.20\%$ & $98.63\%$ \\ \hline
ArcFace & $81.5\%$   & $97.74\%$ & NA    \\   
\hline
\end{tabular}
\caption{TARs for attacks against Facenet and ArcFact at three different FARs.}\label{table:NbNet2ArcFace}
\end{table}
%
%
\section{Theoretical Complements for ECC}\label{app:ECCproofs}
In this section, we provide proofs of propositions that were missing in the main body of this paper due to the space constraints.

\vspace{2mm}
\noindent
\textbf{Proposition 1.} \textit{
Let $\mathcal{C}_\th$ be the set of codewords defined in Definition 1. 
\begin{enumerate}
    \item The minimum angular distance between any two distinct codewords in $\mathcal{C}_\th$ is $\cos^{-1}(1-\frac{1}{\th})$.
    \item The output of the $\mathsf{USample}$ algorithm is uniformly distributed over $\mathcal{C}_\th$.
    \item For any $\tu\in S^{n-1}$, the output $\tc$ of the $\mathsf{Decode}$ algorithm satisfies the following inequality:
$$\langle \tc,\tu\rangle\geq\langle \tc',\tu\rangle\text{ for all }\tc'\in\mathcal{C}_\th.$$
    \item The number of all codewords in $\mathcal{C}_\th$ is $\binom{n}{\th}\cdot2^{\th}$. 
\end{enumerate}
}

\noindent\textit{Proof.}
By Definition 1, the fourth statement is straightforward. Now, we prove the other three statements.
\begin{enumerate}
\item It is sufficient to show that the maximum inner product value is $1-\frac{1}{\th}$ since the cosine function is decreasing between $0$ and $\pi$. By Definition 1, each codeword in $\mathcal{C}_{\th}$ has exactly $\th$ non-zero elements with the same absolute value~$\frac{1}{\sqrt{\th}}$. Thus, it is obvious that two codewords sharing $\th-1$ non-zero positions with the same sign for each entry have the maximum inner product value $\frac{\th-1}{\th}=1-\frac{1}{\th}$.
\item It is sufficient to show that for each $\tc \in \mathcal{C}_{\th}$, the $\mathsf{USample}$ algorithm returns $\tc$ with the probability $1/|\mathcal{C}_\th|$. At Step 1 of the $\mathsf{USample}$ algorithm in Algorithm~\textcolor{red}{4}, the same positions for non-zero elements as those of $\tc$ are selected with the probability $\frac{1}{\binom{n}{\th}}$. At Step 2 of the $\mathsf{USample}$ algorithm, the same sign as those of $\tc$ are selected with the probability $\frac{1}{2^\th}$. Both steps are independent and so the probability that the $\mathsf{Usample}$ algorithm returns $\tc$ is $\frac{1}{2^{\th}\cdot\binom{n}{\th}}$, which is equal to $\frac{1}{|\mathcal{C}_\th|}$ by the fourth statement.
\item Let $\tu=(u_{1},...,u_{n})$ be an element in $S^{n-1}$ and $\tc$ be the output of the $\mathsf{Decode}$ algorithm with input $\tu$. Let $J=\{j_1, \ldots, j_\th\}$ be the set of indices that $\tc$ has non-zero entries.
Then, from the description of the $\mathsf{Decode}$ algorithm, we have
\begin{align*}
\langle \tu,\tc\rangle &= \ u_{j_{1}} \cdot \frac{u_{j_{1}}}{|u_{j_{1}}|\sqrt{\th}}+\cdots+u_{j_{\th}} \cdot \frac{u_{j_{\th}}}{|u_{j_{\th}}|\sqrt{\th}} \\
&=\ \frac{1}{\sqrt{\th}}  \big(|u_{j_{1}}|+\cdots+|u_{j_{\th}}|\big)\ =\ \frac{\sum_{k=1}^{\th}|u_{j_{k}}|}{\sqrt{\th}}.
\end{align*}
Similarly, for any other codeword $\tc'\not=\tc$, the value $\langle\tu,\tc'\rangle$ is equal to the same sum with $J'$ where $J'$ is the set of indices that $\tc'$ has non-zero entries. Thus, the inequality $|u_j|\geq|u_k|$ for $\forall j\in J,\forall k\in[1,n]\setminus J$ given in the description of the $\mathsf{Decode}$ algorithm guarantees that $\langle \tu,\tc\rangle\geq\langle \tu,\tc'\rangle$ for all $\tc'\in\mathcal{C}_\th$ if $J'$ is distinct from $J$.$\hspace{6.25cm} \square$
\end{enumerate}

\vspace{0.2cm}

\noindent\textbf{Proposition 2.} \textit{Given two unit vectors $\tp$ and $\tc$, let $\mathbf{w}=\tc-\tp^{T}\tc\tp$ and $\mathbf{R}_{\theta}=\left[\begin{array}{rr} \cos{\theta} & ~-\sin{\theta} \\ \sin{\theta} & ~\cos{\theta} \end{array}\right]$ where $ \theta=\cos^{-1} (\langle\tp, \tc\rangle)$. Then, the following matrix $\mathbf{R}$ is an orthogonal matrix such that $\mathbf{R}\tp=\tc$:
\begin{align*}
\mathbf{R}=\underbrace{\mathbf{I}-\tp\tp^{T}-\tw\tw^{T}}_{\textrm{projection part}}+\underbrace{[\tp\ \tw]\mathbf{R}_{\theta}[\tp\ \tw]^{T}}_{\textrm{rotation part}}
\end{align*}
where $\mathbf{I}$ is the $n\times n$ identity matrix.
}

\vspace{1mm}

\noindent 
{\textit{Proof.}}    
Since
\begin{eqnarray*}
\mathbf{R}\mathbf{R}^T &=& \left(\mathbf{I}-\tp\tp^{T}-\tw\tw^{T}\right) \left(\mathbf{I}-\tp\tp^{T}-\tw\tw^{T}\right)^T\\
&&+ \left(\mathbf{I}-\tp\tp^{T}-\tw\tw^{T}\right)\left([\tp\ \tw]\mathbf{R}_{\theta}[\tp\ \tw]^{T}\right)^T\\
    &&+\left([\tp\ \tw]\mathbf{R}_{\theta}[\tp\ \tw]^{T}\right)\left(\mathbf{I}-\tp\tp^{T}-\tw\tw^{T}\right)^T \\
    && +[\tp\ \tw]\mathbf{R}_{\theta}[\tp\ \tw]^{T}
\left([\tp\ \tw]\mathbf{R}_{\theta}[\tp\ \tw]^{T}\right)^T\\
&=& \mathbf{I}
\end{eqnarray*}
from the following relations,
\begin{equation*}\begin{aligned}
\left(\mathbf{I}-\tp\tp^{T}-\tw\tw^{T}\right) \left(\mathbf{I}-\tp\tp^{T}-\tw\tw^{T}\right)^T&= \mathbf{I}-\tt\tt^T-\tw\tw^T,\\
[\tp\ \tw]\mathbf{R}_{\theta}[\tp\ \tw]^{T}
\left([\tp\ \tw]\mathbf{R}_{\theta}[\tp\ \tw]^{T}\right)^T &= \tt\tt^T+\tw\tw^T,\\
\left(\mathbf{I}-\tp\tp^{T}-\tw\tw^{T}\right)\left([\tp\ \tw]\mathbf{R}_{\theta}[\tp\ \tw]^{T}\right)^T&= \mathbf{0},~\text{and}
\end{aligned}\end{equation*}
$\mathbf{R}$ is an orthogonal matrix.
Furthermore, since $\tt$ and $\tw$ are orthonormal, we can represent $\tc$ as $\tc=\langle \tc,\tt\rangle \tt +\langle \tc,\tw\rangle \tw= \tt\cos\theta+ \tw\sin\theta=\tc$. 
So, we also have 
\begin{eqnarray*}
\mathbf{R}\tt&&= \left(\mathbf{I}-\tp\tp^{T}-\tw\tw^{T} + [\tp\ \tw]\mathbf{R}_{\theta}[\tp\ \tw]^{T}\right)\tt \\
&& = \tt \cos\theta + \tw \sin\theta=\tc. 
\end{eqnarray*}
Therefore, $\mathbf{R}$ is an orthogonal matrix such that $\mathbf{R}\tt = \tc. \hspace{0.3cm}\square$

\setcounter{equation}{1}
\section{Obstacles for Solving Equation (\textcolor{red}{1}) using a TMTO Strategy}\label{app:tmto}
In this section, we explain how to apply a time-memory-trade-off~(TMTO) strategy to solve Equation (\textcolor{red}{1}) and introduce obstacles that are occurred in the process.
To help the readers' understanding, we first consider the exact version of Equation (\textcolor{red}{1}) that has no error part.
That is, we try to find two codewords~$(\tc_1, \tc_2)$ in $\mathcal{C}_\th$ such that 
\begin{eqnarray}\label{eqn:simplified}
\helper\tc_1 = \tc_2
\end{eqnarray}
for given an orthogonal matrix~$\helper$.
For simplicity, we assume that $\th$ is even, but it is naturally extended to the case that $\th$ is odd.

Let $\mathbf{p}_i$ be the $i$-th column of $\helper$. Then, we can re-write Equation~\eqref{eqn:simplified} as
\begin{eqnarray}\label{eqn:simplified2}
c_{11}\mathbf{p}_1+c_{12}\mathbf{p}_{2}+\cdots+c_{1n}\mathbf{p}_n = \tc_2
\end{eqnarray}
where $\tc_1 = (c_{11}, c_{12}, \cdots, c_{1n})$. 

By using the fact that $\tc_1$ has the exact $\th$ non-zero elements whose absolute values are the same as $\frac{1}{\sqrt{\th}}$, we 
can set two vectors of length $n$, $\mathbf{a}$ and $\mathbf{b}$, such that
they have the exact $\th/2$ non-zero components whose absolute values are the same as $\dfrac{1}{\sqrt{\th}}$, and there is no location that both $\mathbf{a}$ and $\mathbf{b}$ have the non-zero components.
Then, Equation~\eqref{eqn:simplified2} is equivalent to
\begin{eqnarray}\label{eqn:simplified3}
\sum_{\mathbf{a} = (a_1, \cdots, a_n)} a_i\mathbf{p}_i + \sum_{\mathbf{b}=(b_1, \cdots, b_n)} b_j\mathbf{p}_j= \tc_2
\end{eqnarray}
for some $\mathbf{a}$ and $\mathbf{b}$, both of which are determined by $\mathbf{c}_1$. The goal of our TMTO attack is to reduce the searching space of $\tc_1$ at the cost of about square-root size memory and the above equation gives a hint for our purpose. We observe that two terms $\sum_{\mathbf{a} = (a_1, \cdots, a_n)} a_i\mathbf{p}_i$ and $\sum_{\mathbf{b}=(b_1, \cdots, b_n)} b_j\mathbf{p}_j$ in the left hand side of Equation~\eqref{eqn:simplified3} share the same structure and $\tc_2$ in the right hand side is of the special form (e.g., sparse vector consisting of only $0$ and $\pm\frac{1}{\sqrt{\th}}$). When we compute $\sum_{\mathbf{d} = (d_1, \cdots, d_n)} d_i\mathbf{p}_i$ for all possible vectors $\mathbf{d}$ having the exact $\th/2$ non-zero components whose absolute values are $\dfrac{1}{\sqrt{\th}}$ and store them at a table $\mathcal{T}$, both $\sum_{\mathbf{a} = (a_1, \cdots, a_n)} a_i\mathbf{p}_i$ and $\sum_{\mathbf{b}=(b_1, \cdots, b_n)} b_j\mathbf{p}_j$ are elements in $\mathcal{T}$. Then, due to the special form of $\tc_2$, Equation~\eqref{eqn:simplified3} helps us to efficiently search both $\sum_{\mathbf{a} = (a_1, \cdots, a_n)} a_i\mathbf{p}_i$ and $\sum_{\mathbf{b}=(b_1, \cdots, b_n)} b_j\mathbf{p}_j$, not a sequential-manner but at the same time from the table $\mathcal{T}$, which is roughly of square-root size.
%
%


We provide a precise description of our TMTO attack algorithm with detailed analysis.
\begin{enumerate}\itemsep1mm
    \item Generate a table $\mathcal{T}$ that stores all possible pairs of vector~$\mathbf{a}$ and the corresponding vector $\mathbf{q}_i=\sum_{\mathbf{a}} a_i\mathbf{p}_i$
    where $\mathbf{a}$ is a vector of length $n$ that has the exact $\th/2$ non-zero components and their absolute values are $\frac{1}{\sqrt{\th}}$.
    (Note that all possible candidates for $\sum_{\mathbf{a}} a_i\mathbf{p}_i$ are exactly the same as those for $\sum_{\mathbf{b}} b_j\mathbf{p}_j$. So, we can use the same table.)
    
    \item From the table~$\mathcal{T}$, generate a sub-table $\mathcal{T}_i$ which stores pairs of vector $\mathbf{a}$ and the $i$-th component of the corresponding vector $\mathbf{q}_i$ for $1\leq i\leq \ell$. Then, sort tables $\mathcal{T}_i$'s each.
    
    \item For each $\mathbf{q}_i$ stored at $\mathcal{T}$, search a table $\mathcal{T}_i$ to find elements such that the sum of the $i$-th component of $\mathbf{q}_i$ and the value stored at $\mathcal{T}$ is $0$, $\frac{1}{\sqrt{\th}}$, or $-\frac{1}{\sqrt{\th}}$. If the condition holds, we collect the corresponding $(\mathbf{a}, \mathbf{b})$ where $\mathbf{a}$ is the vector corresponding to $\mathbf{q}_i$ in $\mathcal{T}$ and $\mathbf{b}$ is the vector corresponding to the $i$-th component in $\mathcal{T}_i$. Run this step for tables $\mathcal{T}_1, \cdots, \mathcal{T}_\ell$.
    
    \item For all pairs of vectors $(\mathbf{a}, \mathbf{b})$ such that they satisfy the condition in Step~3 for all tables $\mathcal{T}_i$'s and
    there is no location that both $\mathbf{a}$ and $\mathbf{b}$ have non-zero components, check whether 
    \begin{eqnarray*}
    \sum_{\mathbf{a}=(a_1, a_2, \cdots, a_n)} a_i\mathbf{p}_i + \sum_{\mathbf{b}=(b_1, b_2, \cdots, b_n)} b_j\mathbf{p}_j
    \end{eqnarray*}
    is a codeword in $\mathcal{C}_\th$ or not. If it is a codeword, return $\tc_1 = \mathbf{a}+\mathbf{b}$ and $\tc_2 =   \sum_{\mathbf{a}} a_i\mathbf{p}_i + \sum_{\mathbf{b}} b_j\mathbf{p}_j$.
    
\end{enumerate}

Now, we calculate the complexity of the above algorithm. 
Let $N$ be the number of elements stored at $\mathcal{T}$. Then, $N = \binom{n}{\th/2}2^{\th/2}$. Step~1 takes $N$ additions of $\th/2$ $n$-dimensional vectors. Step~2 and Step~3 take $\mathcal{O}(\ell N\log N)$ and $O(\ell\log N)$ comparisons, respectively. Finally, Step~4 takes $c$ additions of $\th$ $n$-dimensional vectors where $c$ is the number of pairs that satisfy the conditions stated in Step~4.
Therefore, if $\ell$ and $c$ is sufficiently small, e.g., both are logarithmic in $N$, then the above algorithm takes a quasi-linear time in $N$.

One may try to check Equation~\eqref{eqn:simplified3} with all candidates after Step~1. We remark that there is no efficient way to execute it since $\tc_2$ is also hidden. That is, the key idea to reduce the complexity of solving by using our method is Steps 2 and 3 which exploits the fact that $\tc_2$ is of the special form that has the exact $\th$ non-zero elements and their absolute values are the same.

Next, we move our attention to solving Equation (\textcolor{red}{1}), which can be regarded as the erroneous version of Equation~\eqref{eqn:simplified}. Similarly to the above, we may apply to the TMTO strategy by mitigating the condition at Step~3 so that it collects the vectors such that the result sum is in some range, not one of the exact values $0, \frac{1}{\sqrt{\th}}$ and $-\frac{1}{\sqrt{\th}}$, to reflect the errors generated by $\mathbf{P}_2\te$ in Equation (\textcolor{red}{1}).
However, in this case, there are still remained many $(\mathbf{a}, \mathbf{b})$'s after Step~3, that is, $c$ is too large in the analysis of the above algorithm.
So, we cannot reduce the complexity of the algorithm as we expect.
%
%

\section{Discussions}
\subsection{Matching Score} 
The verification of IronMask outputs only the binary score, `match'/`no match'.
We would like to note that most biometric cryptographic schemes, which do not use cryptographic encryption schemes, employ cryptographic hash functions. Thus, their verification processes output the binary score as ours. 

Nevertheless, one can indirectly control a threshold for TAR-FAR rate by adjusting parameters of the underlying face recognition system. There is a more direct method for handling this issue by loosening requirements in the matching process: In the registration, a user stores several hashed codewords in close proximity to each other, unlike storing one hashed codeword only in the current registration. Thereafter, in the verification, we may use a modified decoding algorithm that outputs a set of close codewords, instead of only the closest codeword, and check their hashed values with stored ones. From these relaxations, one can control a threshold since those enable to check more approximate matches. Although this direct solution is a plausible candidate, more plentiful analyses and experiments are necessary for resolving this issue completely. 
We leave a detailed analysis as further study.

\subsection{CosFace Experimental Phenomenon}
The performance degradation of CosFace with IronMask is considerable compared to the ArcFace case. (See `CF+I' row in Table~1.) Let us present our interpretation for this experimental
phenomenon. First, we note that we used hyperparameters of ArcFace and CosFace, which were set to get the best performance only, regardless of IronMask. In fact, in order to understand this phenomenon by ourselves, we have evaluated the average cosine values for the same person for both ArcFace and CosFace. As a result, we found that the cosine value of ArcFace is higher than that of CosFace: For example, the average cosine values in ArcFace and CosFace for CMU-MultiPIE dataset are $0.89$ and $0.80$, respectively. We have inferred that IronMask's decoding process relatively well harmonizes with recognition systems with smaller intra-class variation and its boundary lies somewhere between $0.80$ and $0.89$. This causes considerable performance degradation for the CosFace case. To complement this effect, we devise a new centering method using multiple images. In CosFace (and ArcFace), each template is considered as a vector in $S^{n-1}$ with cosine similarity distance and thus simple averaging with the Euclidean distance cannot generate a meaningful center. To overcome this obstacle, we exploited the multiple linear regression to find a center and it improves the intra-class variation: For example, we confirm by experiments that the cosine value of CosFace in the above setting increases from $0.80$ to $0.87$. 
Moreover, as results shown in Table~1, IronMask is well harmonized with both CosFace and ArcFace for several datasets with our new centering method.

\end{document}